\documentclass[10pt,twocolumn,letterpaper]{article}

\usepackage{cvpr}
\usepackage{times}
\usepackage{epsfig}
\usepackage{graphicx}
\usepackage{amsmath}
\usepackage{amssymb}

\usepackage{float}
\usepackage{subfigure}
\usepackage{booktabs}
\usepackage{makecell}
\usepackage{amsfonts}
\usepackage[breaklinks=true,bookmarks=false]{hyperref}

\cvprfinalcopy 

\def\cvprPaperID{****} 

\begin{document}

\title{Student's t-Generative Adversarial Networks}

\author{Jinxuan Sun\\
{\tt\small sunjinxuan1014@gmail.com}\\
\and
Guoqiang Zhong\\
{\tt\small gqzhong@ouc.edu.cn}\\
\and
Yang Chen \\
{\tt\small 374263410@qq.com}\\
\and
Yongbin Liu\\
{\tt\small liuyongbin@stu.ouc.edu.cn}\\
\and
Tao Li\\
{\tt\small 1403371024@qq.com}\\
\and
Zhongwen Guo\\
{\tt\small guozhw@ouc.edu.cn}\\
\and
Department of Computer Science and Technology, Ocean University of China\\
238 Songling Road, Qingdao, China 266100\\
}

\maketitle

\begin{abstract}
Generative Adversarial Networks (GANs) have a great performance in image generation, but they need a large scale of data to train the entire framework, and often result in nonsensical results. We propose a new method referring to conditional GAN, which equipments the latent noise with mixture of Student's t-distribution with attention mechanism in addition to class information. Student's t-distribution has long tails that can provide more diversity to the latent noise. Meanwhile, the discriminator in our model implements two tasks simultaneously, judging whether the images come from the true data distribution, and identifying the class of each generated images. The parameters of the mixture model can be learned along with those of GANs. Moreover, we mathematically prove that any multivariate Student's t-distribution can be obtained by a linear transformation of a normal multivariate Student's t-distribution. Experiments comparing the proposed method with typical GAN, DeliGAN and DCGAN indicate that, our method has a great performance on generating diverse and legible objects with limited data.
\end{abstract}

\section{Introduction}

Generative Adversarial Networks (GANs) \cite{goodfellow2014generative} have been extensively used in image generation. The GAN framework can get nice visual images from simple latent distribution. Typical GAN has two parts respectively called generator and discriminator. The generator G is trained to learn a mapping from a latent space to the data space. Meanwhile, the discriminator D is trained to distinguish between generated and real data. The generator and the discriminator are trained together to compete with each other just like a minimax game. An ideal training result will be obtained when the discriminator cannot tell apart which distribution the data come from. However, there exists model collapse problem \cite{goodfellow2016nips} in GAN. As well as, the generator needs large scale of data to have sufficient capacity \cite{gurumurthy2017deligan} for tacking complexity factors.

Researchers have proposed lots of improvements to GAN to address these problems. \cite{salimans2016improved} modified the discriminator with minibatch, granting the discriminator's internal layers from true date distribution and synthesis data distribution as same as possible. As demonstrated in \cite{warde2016improving}, the authors modified the generator with an additional objective learned by the discriminator using a denoising auto-encoder. Works such as \cite{tolstikhin2017adagan,hoang2018mgan} trained the multi-generator approaches. \cite{nguyen2017dual} used two discriminators with one generater, the two discriminators complete the task of the original discriminator. One discriminator rewards high score for data sampled from true distribution, the other discriminator favoring data from generator distribution. Inspired by conditional variational auto-encoder (CVAE) \cite{DBLP:conf/nips/SohnLY15} and VAE/GAN \cite{DBLP:conf/icml/LarsenSLW16}, CVAE-GAN \cite{bao2017cvae} proposed a frame work that combines a variational auto-encoder with a generative adversarial network under conditional generative process to synthesize fine-grained images. Recent work Generative Adversarial Networks for Diverse and Limited Data (DeLiGAN) \cite{gurumurthy2017deligan} proposed a structure suitable for small and diverse data scenarios, which reparameterizes the latent space as a mixture of Gaussian distribution.
\begin{figure*}[t]
  \centering
  \includegraphics[scale=0.35]{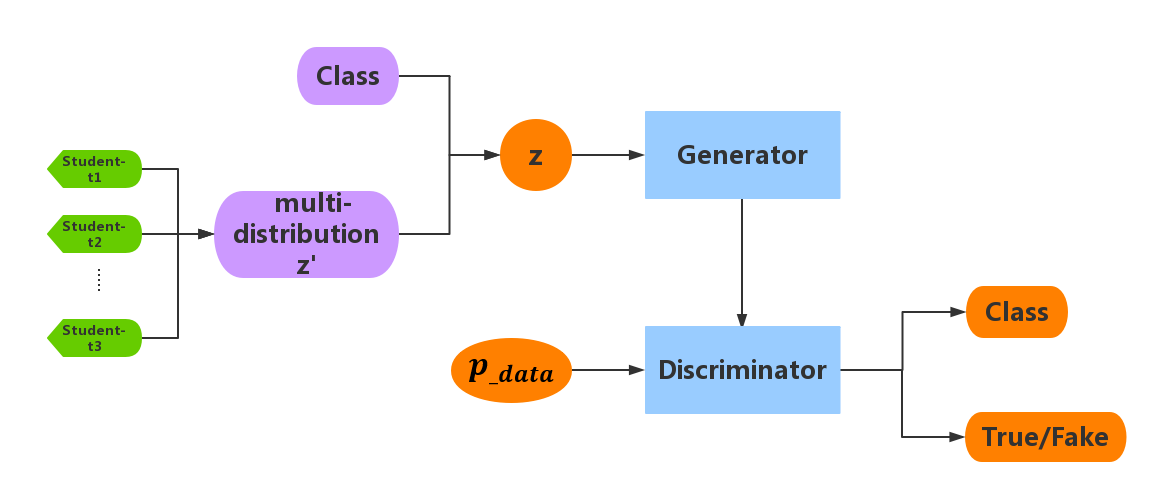}
  \caption{The architecture of tGANs.}\label{CMTGAN}
\end{figure*}

Different from current approaches, the main contribution we make are as four-fold:
\begin{itemize}
\item  We use mixture of $N$ Student's t-distribution (or simply the t-distribution) to simulate the distribution of latent noise, which can provide more diversity than the traditional Gaussian distribution because of its long-tailed effects.
\item  We use attention mechanism to constrain the weights between each of the t-distribution, in this way, we can get a latent noise $z'$ from mixture of the t-distribution.
\item Referring to conditional GAN \cite{mirza2014conditional} concatenating class labels with $z'$, we add an auxiliary classifier to the discriminator, thus our discriminator can not only tell apart real data from generated, but also classify them. By this way, each generated sample achieves an additional class reconstruction error. All of the parameters are learned through the GAN.
\item We mathematically prove that any multivariate Student's t-distribution can be obtained from a standard multivariate Student's t-distribution by a linear transformation.
\end{itemize}

\section{Related Work}

Generative Adversarial Networks (GANs), whose goal is to require an algorithm that can learn the complicated features embodied in image data distributions to generate similar images from scratch, is first proposed by Goodfellow et al \cite{goodfellow2014generative}. A typical GAN has two neural networks, generator G and discriminator D. The generator G is trained to learn a distribution $p_{g}$, which can be matched to real data distribution $p_{data}$. G transforms a latent noise variable from prior distribution $z\sim p_z$ into $G(z)$ which comes from generator distribution $p_{g}$. The discriminator D is trained to distinguish whether $G(z)$ samples from the true data distribution $p_{data}$ (real) or from the synthesized distribution $p_{g}$ (fake). G and D are trained adversarially to compete with each other. The objective function for GAN can be formulated as follows:
\begin{eqnarray}\label{eq:gan}
\begin{aligned}
\min \limits_{G}\max \limits_{D}{V(D,G)}=&E_{x\sim p_{data}}[\log D(x)]\\
&+E_{x\sim p_z}[\log(1-D(G(z)))]
\end{aligned}
\end{eqnarray}

Deep convolutional generative adversarial networks (DCGAN) combined convolutional networks with generative adversarial network for unsupervised training \cite{radford2015unsupervised}, which makes a great improvement for the capability of GAN. CGAN \cite{mirza2014conditional} simply fed the generator and discriminator with extra data to instruct GAN. Multi-agent diverse generative adversarial networks (MGAN) \cite{hoang2018mgan} simultaneously trained a set of generators with parameter sharing, whilst forcing them in different data modes. The recent work called information maximizing
generative adversarial nets (InfoGAN) \cite{chen2016infogan} tried to find an interpretable expression of latent noise. InfoGAN disentangled the latent representation by decomposing the noise to two parts, one is the incompressible noise $z$, the other is an interpretable latent variable $c$. InfoGAN used the maximum mutual information constrain between the latent variable $c$ and generated data, so that $c$ contains interpretable information on the data. Coupled generative adversarial networks (CoGAN) \cite{liu2016coupled} can learn a joint distribution of multi-domain images without requiring tuples of corresponding images in different domains in the training set. The joint distribution is achieved by sharing the parameters of some layers between the generators. DeLiGAN \cite{gurumurthy2017deligan} proposed a GAN-based architecture for diverse and limited training data, which reparameterizes the latent generative space as a mixture of Gaussian model. In \cite{DBLP:journals/corr/Odena16a}, discriminator D was made to predict which of $N+1$ classes ($N$ is the classes number of inputs) to forcing the discriminator network to output class labels. Conditional image synthesis with auxiliary classifier gans (ACGAN) \cite{odena2016conditional} argued GAN with an auxiliary classifier, based on adding label constraints, which improves the quality of high resolution image generation. Previous researches show that GAN with abundant noise can increase the diversity of generated samples, besides, forcing a model to perform additional task is known to improve performance on original task. Motivated by these considerations, we introduce a model that combines both strategies for improve the effects on limited data. That is, we use the mixture of t-distributions to simulate the latent noise, meanwhile, forcing the discriminator to handle the tasks by adding an auxiliary classifier to capture the reconstructed class labels.

Attention mechanism is usually used in neural machine translation tasks. Its principle originating from one important property of human perception that one does not tend to notice a whole visual space in its at once, but focuses attention selectively on parts of the scene to acquire what is needed. Motivated by \cite{bahdanau2014neural,mnih2014recurrent,DBLP:journals/corr/abs-1709-01507}, we utilize the attention mechanism by learn the $\pi_i$ coefficients between each the t-distribution to constrain the weights on each t-distribution, and in this way, a representation of mixture of the t-distribution can be learned. With attention mechanism, we assign different weights to each t-distribution increasing the sensitivity to different t-distribution, so that the noise can super useful ones and suppress less useful ones.

GAN has been successfully applied in many fields. For example, image generation \cite{radford2015unsupervised}, video prediction \cite{liang2017dual}, 3D model generation \cite{wu2016learning}, real image generation based on sketch \cite{ghosh2017multi}, image restoration \cite{bao2017cvae}, superpixel image generation \cite{ledig2016photo}, image-to-image translation \cite{isola2017image}, and text-to-image synthesis \cite{goodfellow2016nips}.

\begin{figure}[h]
  \centering
  \subfigure[Gaussian distribution]{
  \label{(a)Gaussian distribution}
  \includegraphics[scale=0.42]{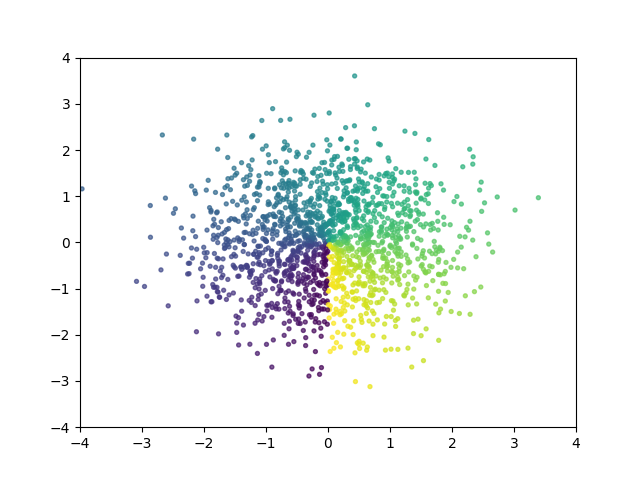}}
  \subfigure[Mixture of Student's t-distribution]{
  \label{(a)Mixture of Student's t-distribution} 
  \includegraphics[scale=0.42]{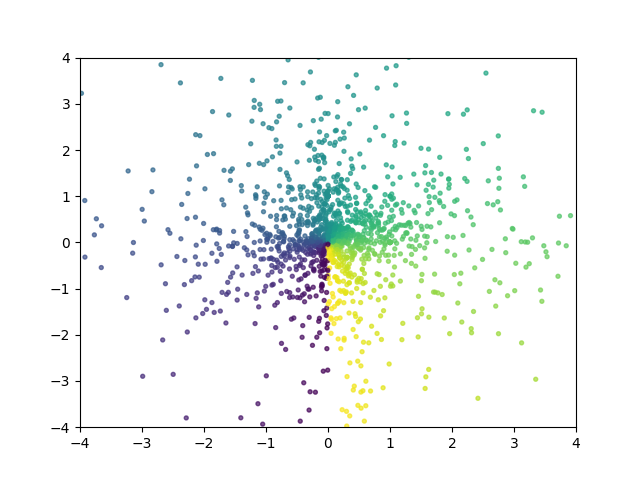}}
  \caption{The top figure shows the Gaussian distribution of the original noise vectors while the bottom one is the mixture of Student's t-distribution noise of our model. We can see that the points in the bottom figure are more widely distributed.}
  \label{noise-distribution}
\end{figure}
\begin{figure*}[t]
  \centering
  \includegraphics[scale=0.25]{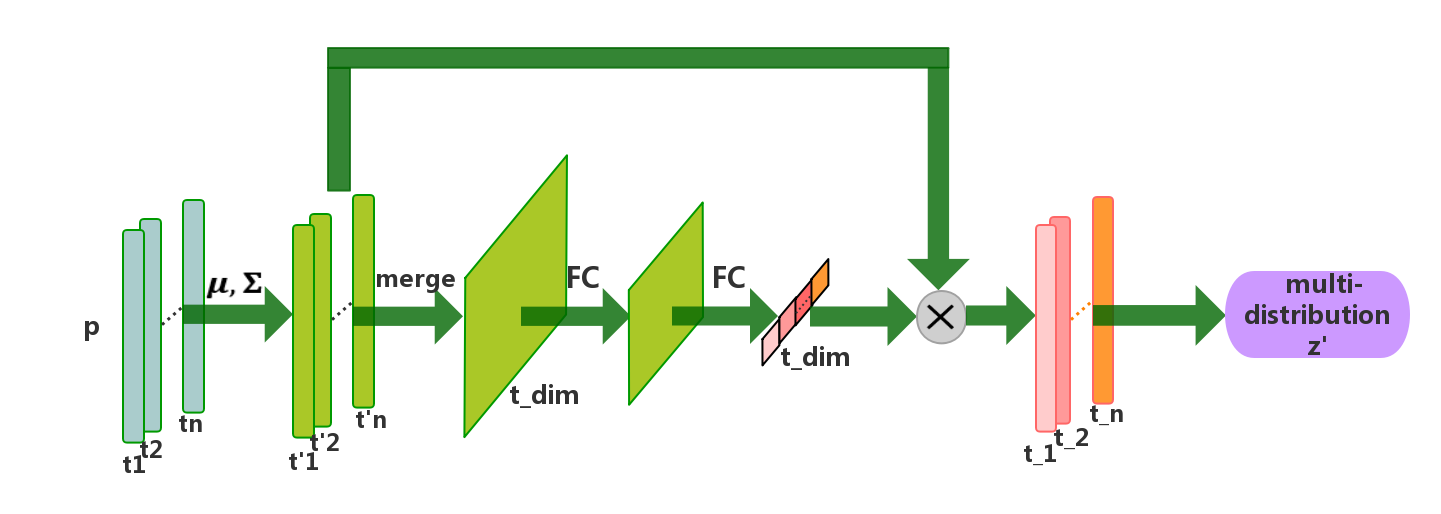}
  \caption{Attention mechanism we used to weights between the t-distribution components.}\label{attention}
\end{figure*}

\section{Student's t-distribution Generative Adversarial Networks (tGANs)}

To generate diversity examples with limited data, we propose tGANs which can generate diverse object, the auxiliary classifier in discriminator can also help to avoids the collapse of the model.

 Our idea is to use the mixture of Student's t-distribution combine with conditional information as latent noise, not a single distribution in the standard GAN \cite{goodfellow2014generative} nor mixture of Gaussian distribution in DeLiGAN \cite{gurumurthy2017deligan}. Different from MGAN \cite{ghosh2017multi} with multiple generator, our model provide more diversity to the latent noise that we need lower computational complexity. Compare to CGAN \cite{mirza2014conditional}, we can provide more diverse samples on certain condition. Figure \ref{noise-distribution} illustrates the noise of Gaussion distribution and our mixture of Student's t-distribution with attention mechanism. We use colorful points to show the space of noise. As we can see, the mixture of Student's t distribution has a wider range with higher complexity to better cover the latent noise space. In this paper, we combine attention mechanism with the mixture of t-distribution to constraint the latent noise $P_z$, enlarging the divergence of $z$ with semantic features of data. The auxiliary classifier in discriminator gives out class information to constraint generated examples to certain category, avoiding generating meaningless images and speed up convergence.
\subsection{Theoretical Analysis}
In this part we mathematically prove that the linear transforms of a multivariate t-distribution are still a multivariate t-distribution ${\boldsymbol t}({\boldsymbol \mu ,\boldsymbol \Sigma,\boldsymbol \nu})$. The density of $\boldsymbol x \in {\mathbb{R}^p}$ is defined as follow:
\begin{eqnarray}\label{eq:multt}
\begin{aligned}
f(\boldsymbol x)
=&\frac{\Gamma(\frac{\boldsymbol\nu+p}{2})}{\Gamma(\frac{\boldsymbol \nu}{2})\boldsymbol\nu^\frac{p}{2}\pi^\frac{p}{2}\vert\boldsymbol
\Sigma\vert^\frac{1}{2}}\\
&[1+\frac{1}{\boldsymbol\nu}(\boldsymbol{x}-\boldsymbol\mu)^T\boldsymbol\Sigma^{-1}(\boldsymbol{x}-\boldsymbol\mu)]
^{-\frac{\boldsymbol\nu+p}{2}}
\end{aligned}
\end{eqnarray}
where $p$ is the dimension of variant, $\boldsymbol\Sigma$ is a $p\times p$ positive-definited real matrix (we assume that the various distributions are irrelevant, so the $\boldsymbol\Sigma$ can be a diagonal matrix which has $\sigma_{11}^2$, $\sigma_{22}^2$,... , $\sigma_{pp}^2$ on the diagonal ), $\boldsymbol\mu = [\mu_1, \dots, \mu_p]^T$ is the mean value of each distribution , $\boldsymbol\nu=[\nu_1, \dots, \nu_p]^T$ is the degrees of freedom, $\nu_1 =\nu_2=\dots=\nu_p=\nu$ so we simply use $\nu$ to express $\boldsymbol\nu$.

We linearly transform $\boldsymbol x=(x_1,x_2,...,x_p)^T$ to $\boldsymbol y=(y_1,y_2,...,y_p)^T$ by the function $y_i$,
\begin{equation}
y_i=g_i(x_i)=\frac {x_i-\mu_i}{\sigma_{ii}}
\end{equation}
 $x_i$ can be shown as the inverse function of $g_i(x_i)$,
\begin{equation}
h_i(y_i)=\sigma_{ii}y_i+\mu_i
\end{equation}
The possible density distribution of $\boldsymbol y$ as:
\begin{equation}\label{eq:td}
\begin{aligned}
&\quad f{y_1},...,{y_p}({y_1},...,{y_p}) \\
&=f{x_1},...,{x_p}({h_1}({y_1}),...,h_p({y_p}))
 \times \left| {J({y_1},...,{y_p})} \right|\\
 &= f{x_1},...,{x_p}({\sigma_{11}y_1+\mu_1},...,\sigma_{pp}y_p+\mu_p)\times |\det {D_y}|\\
 &= \frac{\Gamma(\frac{\nu+p}{2})|\det {D_y}|}
 {\Gamma(\frac{\nu}{2})\nu^\frac{p}{2}\pi^\frac{p}{2}\vert\boldsymbol\Sigma\vert^\frac{1}{2}}\\
&\quad[1+\frac{1}{\nu}(\sigma_{11}y_1,...,\sigma_{pp}y_p)^T\boldsymbol\Sigma^{-1}
(\sigma_{11}y_1,...,\sigma_{pp}y_p]^{-\frac{\nu+p}{2}}\\
 &=\frac{\Gamma(\frac{\nu+p}{2})|\boldsymbol\Sigma^\frac{1}{2}|}
 {\Gamma(\frac{\nu}{2})\nu^\frac{p}{2}\pi^\frac{p}{2}\vert\boldsymbol\Sigma\vert^\frac{1}{2}}
[1+\frac{1}{\nu}(\boldsymbol\Sigma^\frac{1}{2}\boldsymbol y)^T\boldsymbol\Sigma^{-1}
(\boldsymbol\Sigma^\frac{1}{2}\boldsymbol y)]^{-\frac{\nu+p}{2}}\\
 &=\frac{\Gamma(\frac{\nu+p}{2})}
 {\Gamma(\frac{\nu}{2})\nu^\frac{p}{2}\pi^\frac{p}{2}}
[1+\frac{1}{\nu}\boldsymbol y^T\boldsymbol y]^{-\frac{\nu+p}{2}}\\
 \end{aligned}
 \end{equation}
 we have known from the other works, that
 \begin{equation}\label{eq:3}
\begin{array}{l}
|J(y_1,y_2,...,y_k)| =|\det {D_y}|
\end{array}
\end{equation}
 As Eq.\ref{eq:td}, we can find that any multivariate of t-distribution can be transformed to a standard t-distribution by a linear transformation. In others words, we prove that any multivariate of t-distribution can be obtained from a standard t-distribution.

 The Jacobian of the transformations can be shown as follows:
\begin{align}
\label{eq:0} {D_x} &=
\left[ {\begin{array}{*{20}{c}}
{\frac{{\partial {y_1}}}{{\partial {x_1}}}}&{\frac{{\partial {y_1}}}{{\partial {x_2}}}}&{ \cdot  \cdot  \cdot }&{\frac{{\partial {y_1}}}{{\partial {x_p}}}}\\
{\frac{{\partial {y_2}}}{{\partial {x_1}}}}&{\frac{{\partial {y_2}}}{{\partial {x_2}}}}& \ldots &{\frac{{\partial {y_2}}}{{\partial {x_p}}}}\\
 \vdots & \vdots & \ddots & \vdots \\
{\frac{{\partial {x_p}}}{{\partial {x_1}}}}&{\frac{{\partial {y_k}}}{{\partial {x_2}}}}& \ldots &{\frac{{\partial {y_k}}}{{\partial {x_p}}}}
\end{array}} \right]\notag\\
&=\left[ {\begin{array}{*{20}{c}}
{{\sigma _{11}^{-1}}}&{}&{}&{}\\
{}&{{\sigma _{22}^{-1}}}&{}&{}\\
{}&{}&\ddots&{}\\
{}&{}&{}&{{\sigma _{pp}^{-1}}}
\end{array}} \right]
=\boldsymbol\Sigma^{-\frac {1}{2}}\\
\label{eq:1} {D_y} &=
\left[ {\begin{array}{*{20}{c}}
{\frac{{\partial {x_1}}}{{\partial {y_1}}}}&{\frac{{\partial {x_1}}}{{\partial {y_2}}}}&{ \cdot  \cdot  \cdot }&{\frac{{\partial {x_1}}}{{\partial {y_p}}}}\\
{\frac{{\partial {x_2}}}{{\partial {y_1}}}}&{\frac{{\partial {x_2}}}{{\partial {y_2}}}}& \ldots &{\frac{{\partial {x_2}}}{{\partial {y_p}}}}\\
 \vdots & \vdots & \ddots & \vdots \\
{\frac{{\partial {x_p}}}{{\partial {y_1}}}}&{\frac{{\partial {x_p}}}{{\partial {y_2}}}}& \ldots &{\frac{{\partial {x_p}}}{{\partial {y_p}}}}
\end{array}} \right]\notag\\
&=\left[ {\begin{array}{*{20}{c}}
{{\sigma _{11}}}&{}&{}&{}\\
{}&{{\sigma _{22}}}&{}&{}\\
{}&{}&\ddots&{}\\
{}&{}&{}&{{\sigma _{pp}}}
\end{array}} \right]
=\boldsymbol\Sigma^{\frac {1}{2}}
\end{align}
\subsection{tGANs}
In GAN training, we attempt to learn a mapping from a simple latent space to the complicated data distribution. The generator G needs to have sufficient capacity to capture the relationship between the simple distribution with the complicated distribution. Instead of using multiple generators or increasing the model depth which will cost a lot of computational complexity, we propose a variant of the GAN architecture which we call Student's t Generative Adversial Networks (tGANs).
\subsubsection{Mixture of Students' t-distribution}
In tGANs, we reparameter the noise $z'$ by the mixture of t-distribution ${\boldsymbol t_i}({\boldsymbol\mu _i},{\boldsymbol\Sigma _i},{\nu})$. For simplicity, we use the same degree of freedom ${\nu}$ for each t-distribution. We randomly sample $N$ of t-distribution from standard t-distribution $T(0,1,\nu)$, and reparameterize each t-distribution with ${\boldsymbol\mu_i}$ and ${\boldsymbol\sigma_i = [\sigma_{i1}, \sigma_{i2},\dots ,\sigma_{ip}]}$, $p$ is the dimension of each t-distribution.
\begin{eqnarray}\label{eq:eacht}
{\boldsymbol t_i}={\boldsymbol\mu _i}+{ \boldsymbol\sigma _i}\epsilon  \ \ \ \ \ \ \  \epsilon\in T(0,1,\nu)
\end{eqnarray}
Therefore, obtaining a latent space sample translates to sampling $\epsilon\sim T(0,1,\nu)$, and calculate $t_i$ according to Eq.\ref{eq:eacht}. The latent space $p_{z'}(z') $ we get:
\begin{eqnarray}\label{eq:pz}
{p_{z'}}(z') = \sum\limits_{i = 1}^N {{\pi _i}{\boldsymbol t_i}({\boldsymbol\mu _i},{\boldsymbol\Sigma _i},{\nu})}
\end{eqnarray}
where $\pi _i$ is the coefficients of the mixture distribution which can be learned by attention mechanism. To obtain a sample from ${p_{z'}}(z')$ distribution, we randomly sample from the distribution.

\subsubsection{Conditional Noise and Attention Mechanism}
The coefficients $\pi _i$ of the mixture distribution is learned by attention mechanism as Figure \ref{attention}. We merge $N$ number of t-distributions with $p$ dimensions into a vector, following with two fully connected layers to map the vector into the dimension of $N$ (the number of t-distribution). We use the final vector as weight $\boldsymbol\pi$ for each t-distribution. Since the output is produced by a summation through all of t-distribution, the dependencies of each t-distribution are implicitly embedded in $\boldsymbol\pi$. Our goal is to ensure that the attention network is able to increase the sensitivity to different t-distribution so that the noise can super useful ones and suppress less useful ones. Here,we train $\pi _i$ simultaneously along with the whole networks.

Based on CGAN, every generated sample has a corresponding class label $c$ in addition to the noise $z'$ we get above. Generator uses both to generate images $G(z,c)$.
\subsubsection{Discriminator with Auxiliary Classifier}
The discriminator of our model can achieve tasks on two aspects, telling apart generated data from real data and classifying generated samples which class they belong to. We implement the tasks by adding an auxiliary classifier to the discriminator. Different from adding another classifier with increasing lots of calculation cost, we use first three layers of the discriminator as the input of the auxiliary classifier. The auxiliary classifier has 3 layers of convolution, kernel size $3\times3$, stride 1, with dropout and softmax. The object for the auxiliary classifier is to optimize the loss function,
\begin{align}
\label{eq:lsc} L_{C}=&E_{x\sim p_{data}}[\log C(x)]\notag\\
 &+E_{x\sim p_{z,c}}[\log C(G(z,c))]
\end{align}
The classification part can add more insight to the latent variables, representing and disentangling potentially manipulation the generated images. This modification produces excellent results and appears to stabilize training.

In traditional GAN, the generator G and discriminator D try to optimize the following equation,
\begin{align}
\label{eq:lsgt} G_{loss}=&E_{x\sim p_{z}}[\log(1-D(G(z)))]\\
\label{eq:lsdt} D_{loss}=&E_{x\sim p_{data}}[\log D(x)]\notag\\
 &+E_{x\sim p_{z}}[\log(1-D(G(z)))]
\end{align}

In tGANs, we train the generator G and the discriminator D by alternatively minimizing $L_{G}$ in Eq.\ref{eq:lsg} and maximizing $L_{D}$ in Eq.\ref{eq:lsd},
\begin{align}
\label{eq:lsg} L_{G}=&E_{x\sim p_{z,c}}[\log(1-D(G(z,c)))]+\alpha L_{C}\\
\label{eq:lsd} L_{D}=&E_{x\sim p_{data}}[\log D(x)]\notag\\
 &+E_{x\sim p_{z,c}}[\log(1-D(G(z,c)))]+\alpha L_{C}
\end{align}
Figure\ref{CMTGAN} illustrates the general architecture of our proposed tGANs.

Structurally, our model is based on existing models, borrow the mixture distribution from DeLiGAN \cite{gurumurthy2017deligan} and the conditional side information from ACGAN \cite{odena2016conditional}. However, the modification to the latent noise $z$ with attention mechanism can get efficient improvement to produce excellent diversity results on limited data, the auxiliary classifier in discriminator appears to stabilize training. Moreover, we mathematically prove that any mixture of t-distribution can be obtained by transforming from the standard t-distribution.
\section{Experiments}

In this section, we conduct experiment on MNIST \cite{lecun1998gradient}, Fashion MNIST \cite{xiao2017/online} and CIFAR-10 datasets \cite{krizhevsky2009learning}. We compare to previous work typical GAN, DCGAN and DeLiGAN, and we use inception score to evaluate the generated effect. The experimental results demonstrate its efficacy and scalability of stable training and provide diverse images.

For our tGANs framework, the choice of $N$ and $p$, the number of t-distribution and the dimensions of each t-distribution components, are made empirically more complicated data distributions requires larger $N$ and $p$. Dimensions of $z$ equals to $N\times p$, increasing $z_{dim}$ also increasing memory requirements. Experiments indicate that blindly increasing $z_{dim}$ has no contribution on model capacity. We use a $N$ between 5 and 50, a $p$ between 10 and 25 for our experiments. Besides, we use $\alpha$ as 1 for simplicity.
\subsection{Experiments on the MNIST Dataset}
MNIST is a dataset of handwritten digits from 0 to 9, which has a training set of 60,000 examples, and a test set of 10,000 examples. It is a subset of a larger set available from NIST. The digits have been size-normalized as $28\times28$ and centered in a fixed-size image. In our experiment, only 500 images are taken into count, each of them is randomly sampled from the dataset, keeping balance of per digit. For MNIST, the generator network has 2 layers of fully connected layer followed by the other 2 deconvolution layers, the discriminator network has 3 convolutional layers followed by 2 fully connected layers to distinguish whether the input images came from true data distribution, another 2 fully connected layers with softmax to identify the class of generated images.

\begin{figure}[h]
  \centering
  \includegraphics[scale=0.4]{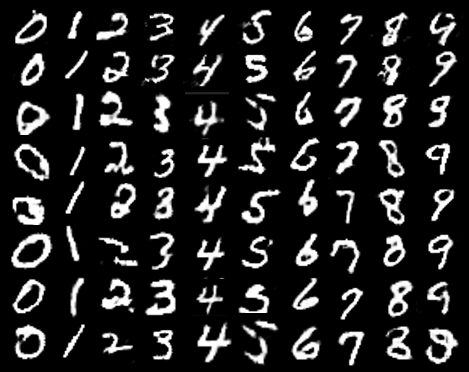}
  \caption{The 0 to 9 digits generated by our model trained using the MNIST handwritten digits.}\label{cmcmnist09}
\end{figure}

In Figure \ref{cmcmnist09}, we show $0$-$9$ digits samples generated by our models. Each column demonstrate a class of the digit from $0$ to $9$. We can see from that, the digits in each column has variate to the other. For digit $1$, we can easily observe that they have different slope angles. For digit $2$, there are different types depend on the shape of tails, some are a straight line, some have a circle, and some are like a wavy. As for digit $9$, there are both slope angles and tails different for the digits.

\begin{figure}[h]
\centering
\subfigure[tGANs]{
\label{(a)tGANs} 
\includegraphics[width=1.5in, height=1.5in]{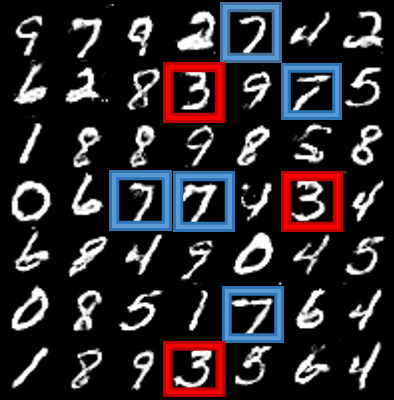}}
\quad
\subfigure[GAN]{
\label{(b)GAN} 
\includegraphics[width=1.5in, height=1.5in]{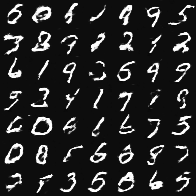}}
\quad
\subfigure[DeLiGAN]{
\label{(c)DeLiGAN} 
\includegraphics[width=1.5in, height=1.5in]{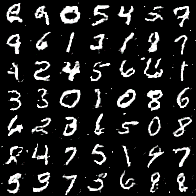}}
\quad
\subfigure[DCGAN]{
\label{(d)DCGAN} 
\includegraphics[width=1.5in, height=1.5in]{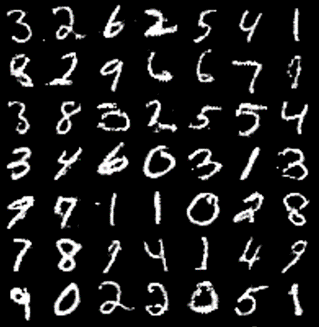}}
\caption{Comparing tGANs with typical GAN, DeLiGAN and DCGAN}
\label{tGANs compare with typical GAN, DeLiGAN and DCGAN} 
\end{figure}

In Figure \ref{tGANs compare with typical GAN, DeLiGAN and DCGAN}, we show samples generated by different methods as tGANs, typical GAN, DeLiGAN and DCGAN. For each model, we present the object we generate with same steps. tGANs, GAN and DeLiGAN used the limited dataset which only has 500 images, but DCGAN used the total 60,000 dataset. DCGAN need a large scale of data to train the entire framework, with limited data in limited step it cannot give out satisfied result. The samples produced by our model (Figure \ref{(a)tGANs}, top-left) is much more clearly than typical GAN (Figure \ref{(b)GAN}, top-right). GAN gets more meaningless pattern which cannot recognize as digits. DeLiGAN (Figure \ref{(c)DeLiGAN}, bottom-left) generates numbers with diversity, but it is lack of supervised information that lead to deformed unlikeness digits. DCGAN has a great performance to generate samples identical to one other (Figure \ref{(d)DCGAN}, bottom-right), nevertheless our tGANs can provide more style information to the digits. As the numbers of $3$ which are circled by red rectangles, the degree of inclination of the digits is different, besides, the lower part has different curvature. There are more diversity in blue rectangles for such a single digit $7$. Firstly, the horizontal stroke has different lengths, secondly, on the beginning of the horizontal, there can be a vertical bar, thirdly, the slash has different degree of inclination.

\begin{table}[h]
\centering
\scalebox{0.9}[0.9]{
\begin{tabular}{ccccc}
  \hline
  \thead{Inception Score}& \thead{tGANs} &\thead{ GAN} &\thead{ DeLiGAN} &\thead{ DCGAN} \\
  \hline
  \thead{ Mean }&\thead{ 1.8039} & \thead{1.1135} &\thead{ 1.3886 }& \thead{1.6528} \\
  \hline
  \thead{Standard} & \thead{0.0063 }& \thead{0.0950} &\thead{ 0.0160} & \thead{0.1140 }\\
  \hline
\end{tabular}}
  \caption{Comparing inception score value for typical GAN, DeLiGAN and DCGAN with our model. Model with larger scores means that the generated digits are better.}
  \label{mnist-incep}
\end{table}
\subsection{Experiments on the Fashion MNIST Dataset}
Fashion MNIST, similar to MNIST, which has a training set of 60,000 $28\times28$ grayscale images, and a test set of 10,000 $28\times28$ grayscale images. Different from MNIST, the 10 categories of Fashion MNIST are taken from fashion objects, like: T-shits, trouser, pullover, dress, coat, sandal, shirt, sneaker, bag, ankle boot. For the experiment involving Fashion MNIST, we adopt the same architecture as we used in MNIST. The generator network has 2 layers of fully connected layer followed by the other 2 deconvolution layers, the discriminator network has 3 convolutional layers followed by 2 fully connected layers to distinguish if the input images came from true data distribution, another 2 fully connected layers with softmax to identify the class of the generated images.

\begin{figure}[h]
\centering
\subfigure[tGANs]{
\label{(a)f-tGANs} 
\includegraphics[width=1.5in, height=1.5in]{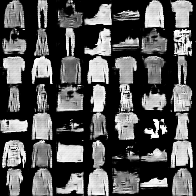}}
\quad
\subfigure[GAN]{
\label{(b)f-GAN} 
\includegraphics[width=1.5in, height=1.5in]{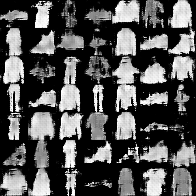}}
\quad
\subfigure[DeLiGAN]{
\label{(c)f-DeLiGAN} 
\includegraphics[width=1.5in, height=1.5in]{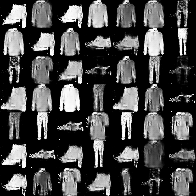}}
\quad
\subfigure[DCGAN]{
\label{(d)f-DCGAN} 
\includegraphics[width=1.5in, height=1.5in]{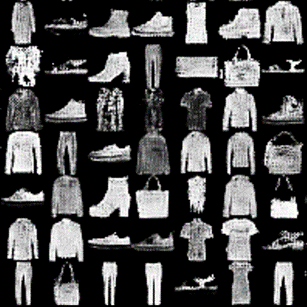}}
\caption{Comparing tGANs with typical GAN, DeLiGAN and DCGAN.}
\label{tGANs, typical GAN, DeLiGAN and DCGAN on Fashion-MNSIT} 
\end{figure}

In Figure \ref{tGANs, typical GAN, DeLiGAN and DCGAN on Fashion-MNSIT}, we show Fashion MNIST samples generated by different method as tGANs,typical GAN, DeLiGAN and DCGAN. For each model, we present the objects we generate under the same conditions. The samples produced by our model (Figure \ref{(a)f-tGANs}, top-left) are much more clearly than GAN (Figure \ref{(b)f-GAN}, top-right). DCGAN has the best visual performance (Figure \ref{(d)f-DCGAN}, bottom-right). To further explore the diversity of generated results, we evaluate each class of them using inception score. The values of each class is demonstrated in figure \ref{fashion-incep}.

\begin{figure}[h]
  \centering
  \includegraphics[scale=0.55]{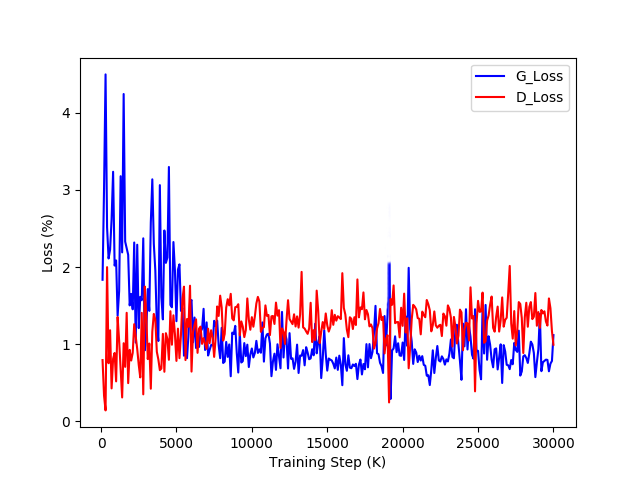}
  \caption{The $G_{loss}$ and $D_{loss}$ values changing curve of tGANs.}\label{dgloss}
\end{figure}

In order to track the training process, we printed the $D_{loss}$ and $G_{loss}$ every 300 steps. Figure \ref{dgloss} display the value changing cure of tGANs. Note that, at beginning $G_{loss}$ is high and $D_{loss}$ is close to $0$. As the number of step increasing, $D_{loss}$ gradually stabilizes at around $1.3$. As we know, the generator is well trained when the discriminator cannot tell apart generated from real data, at this time, $\log D(x)$ and $\log(1-D(G(z)))$ both approximate $0.5$, and $D_{loss}$ in Eq.\ref{eq:lsdt} approximate $2log2$. After 7500 steps $D_{loss}$ of our model is stabilized around $1.3$ which is approaches to $2log2$, infers that our modification on GANs can stabilize training and accelerate convergence.
\begin{table*}[t]
\centering
\scalebox{0.75}[0.75]{
\begin{tabular}{ccccccccccc}
  \hline
  \thead{Inception Score} &\thead{T-shirt/top}  &\thead{Trouser} &\thead{Pullover }  &\thead{ Dress} &\thead{ Coat} &\thead{ Sandal} &\thead{ Shirt} &\thead{Sneaker } &\thead{Bag } &\thead{ Ankle boot}\\
  \hline
  \thead{tGANs }&\thead{ $1.48\pm0.09$} &\thead{ $1.28\pm0.09$ }& \thead{$\boldsymbol{1.61\pm0.11} $} &\thead{$ \boldsymbol{1.67\pm0.07 }$ } &\thead{$ \boldsymbol{1.89\pm0.16 }$ }& \thead{$ 1.04\pm0.07$}&\thead{$\boldsymbol{1.81\pm 0.21 }$ } &\thead{ $\boldsymbol{ 1.15\pm 0.12} $}& \thead{$ \boldsymbol{1.38\pm 0.06 }$}& \thead{$ 1.27\pm0.14$}\\
  \hline
  \thead{DeLiGAN} &\thead{ $1.38\pm0.11 $} &\thead{ $1.28\pm0.10 $ }& \thead{$1.42\pm0.14 $} &\thead{  $1.53\pm0.09$} &\thead{ $1.75\pm0.07$ }& \thead{$1.03\pm0.05$}&\thead{$1.65\pm0.17 $ } &\thead{$1.13\pm0.11 $  }& \thead{$1.25\pm0.10 $}& \thead{$1.22\pm0.05 $}\\
  \hline
  \thead{DCGAN} &\thead{$1.51\pm0.14 $} &\thead{ $1.34\pm0.09 $ }& \thead{$1.58\pm0.17 $} &\thead{ $1.66\pm0.11$} &\thead{ $1.88\pm0.15$ }& \thead{$1.09\pm0.10$}&\thead{ $1.80\pm0.20 $} &\thead{ $1.12\pm0.05 $ }& \thead{$1.36\pm0.11 $}& \thead{$1.29\pm0.17 $}\\
  \hline
\end{tabular}}
  \caption{Comparing the generated samples' inception score of tGANs, DeLiGAN and DCGAN on the Fashion MNIST dataset. Model with larger scores means that the generated images are better.}
  \label{fashion-incep}
\end{table*}

\subsection{Experiments on the CIFAR-10 Dataset}
CIFAR-10 is a dataset which contains 60,000 color images in 10 classes, each class has 6000 images. The images have been size-normalized as $32\times32$, associated with a label from airplane, automobile, bird, cat, deer, dog, frog, horse, ship and truck. We conduct experiments on a limited data with 5000 images randomly selected from CIFAR-10 dataset. For the experiment involving CIFAR dataset, we adopt the architecture that the generator network has a fully connected layer followed by 3 deconvolution layers with batch normalization after each layer. The first 3 layers in discriminator networks are convolutional layers with dropout and batch normalization, the 4th and 5th layers are fully connected layers which tell apart generated from real data, like most discriminators, the 6th to 11th layers are auxiliary classifiers.
\begin{table}[h]
\centering
\begin{tabular}{ccccc}
  \hline
  \thead{Inception Score}& \thead{tGANs}  &\thead{ DeLiGAN} &\thead{ DCGAN} \\
  \hline
  \thead{ Mean }&\thead{ 2.4352} &\thead{ 1.6235 }& \thead{2.3911} \\
  \hline
  \thead{Standard} & \thead{0.2343 } &\thead{ 0.3188} & \thead{0.3370 }\\
  \hline
\end{tabular}
  \caption{Comparing inception score values for tGANs, DeLiGAN and DCGAN. Model with larger scores means that the generate images are better.}
  \label{cifar-incep}
\end{table}
\begin{figure}[h]
  \centering
  \includegraphics[scale=0.35]{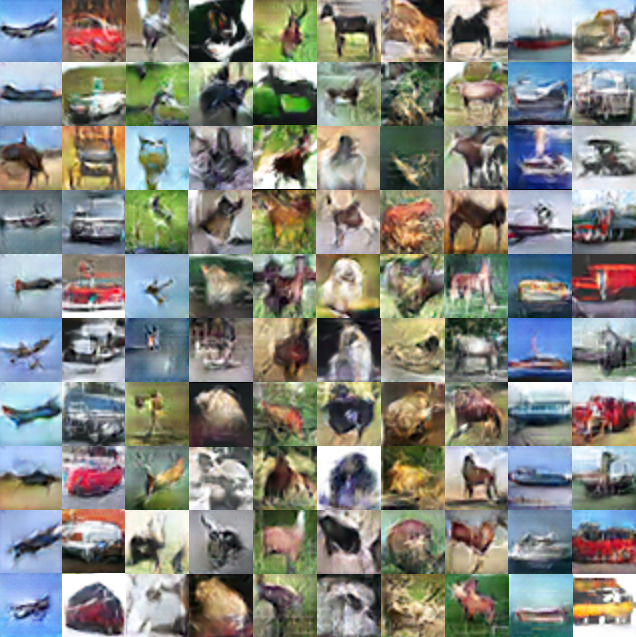}
  \caption{These images are samples from the set of generated images using CIFAR-10 as input data. The labels from left to right are airplane, automobile, bird, cat, deer, dog, frog, horse, ship, truck.}\label{cifar}
\end{figure}

Figure \ref{cifar} shows samples generated by our model in different classes, which demonstrate the generation ability of our model on tiny images. The inception score values for different models are displayed in Table \ref{cifar-incep}. Compared to DeliGAN and DCGAN, our model has a higher mean values, indicating that our generated images are more diverse.

\section{Conclusion}

In this paper, we propose Student's t-distribution Generative Adversarial Networks (tGANs) which can generate diverse objects with limited data. The proposed method uses the mixture of t-distribution with weighted by attention mechanism combined with conditional information to reparameterize the latent noise space. Our argument to discriminator that adding an auxiliary classifier provides more insight to the latent variables represent. Extensive quantitative and qualitative results demonstrate the effectiveness of our propose method. Compared with existing models as DCGAN and DeLiGAN, our method generates diverse samples stably.

{\small
\bibliographystyle{ieee}
\bibliography{egbib}
}

\end{document}